\documentclass[10pt,twocolumn,letterpaper]{article}

\usepackage{authblk}
\usepackage[pagenumbers]{cvpr} 

\usepackage[accsupp]{axessibility}  
\usepackage{graphicx}
\usepackage{amsmath}
\usepackage{amssymb}
\usepackage{paralist}
\usepackage{booktabs}
\usepackage[inline]{enumitem}
\usepackage{multirow}
\usepackage{placeins}
\usepackage{dblfloatfix}
\usepackage{caption}
\usepackage{subcaption}
\makeatletter
\@namedef{ver@everyshi.sty}{}
\makeatother
\usepackage{tikz}

\usepackage[pagebackref,breaklinks,colorlinks]{hyperref}
\makeatletter

\renewcommand\AB@affilsepx{\quad \protect\Affilfont}
\makeatother
\newcommand{\paragraphNoSpace}[1]{{\bf #1 \hspace{0.05cm}}}

\usepackage[capitalize]{cleveref}
\crefname{section}{Sec.}{Secs.}
\Crefname{section}{Section}{Sections}
\Crefname{table}{Table}{Tables}
\crefname{table}{Tab.}{Tabs.}

\def\cvprPaperID{7537} 
\def\confName{CVPR}
\def\confYear{2022}

\begin{document}

\title{Dense Depth Priors for Neural Radiance Fields from Sparse Input Views}
\author[1]{\vspace{-0.2cm}Barbara Roessle}
\author[2]{Jonathan T. Barron}
\author[2]{Ben Mildenhall}
\author[2]{Pratul P. Srinivasan}
\author[1]{Matthias Nie\ss{}ner}
\affil[1]{Technical University of Munich}
\affil[2]{Google Research}

\twocolumn[{%
\renewcommand\twocolumn[1][]{#1}%
\maketitle
\begin{center}
    \centering
    \captionsetup{type=figure}
    \vspace{-0.6cm}
    \includegraphics[width=\textwidth,trim={0.1cm 4.6cm 0 4.7cm 0.1cm},clip]{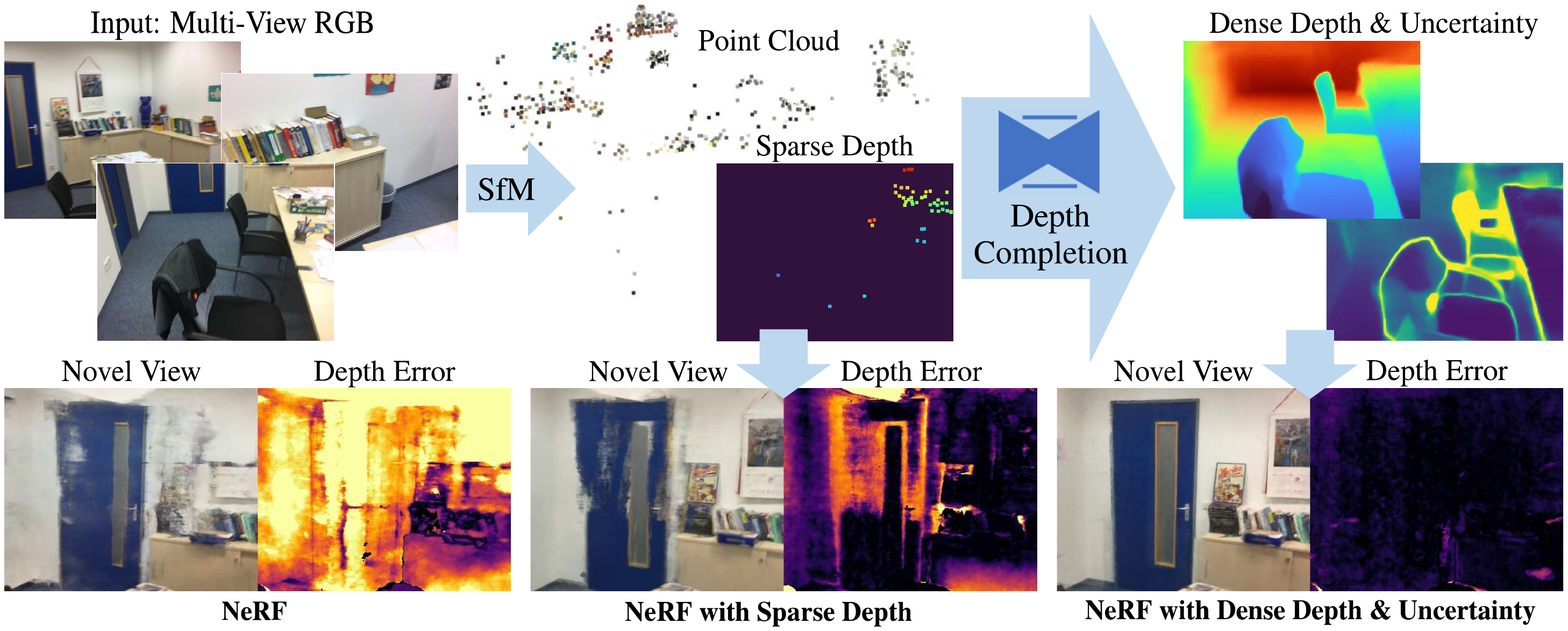}
    \captionof{figure}{We leverage dense depth priors for recovering neural radiance fields of complete rooms when only a handful of input images are available. To this end, we first utilize the sparse point cloud reconstructions from SfM preprocessing, which we feed into a depth completion network. 
    We then impose these depth estimates as constraints to the NeRF optimization according to the estimated uncertainty.
    This facilitates novel view synthesis results at significantly higher image quality and lower depth error compared to NeRF.
    }
    \label{fig:teaser}
\end{center}%
}]

\begin{abstract}
Neural radiance fields (NeRF) encode a scene into a neural representation that enables photo-realistic rendering of novel views. 
However, a successful reconstruction from RGB images requires a large number of input views taken under static conditions --- typically up to a few hundred images for room-size scenes. 
Our method aims to synthesize novel views of whole rooms from an order of magnitude fewer images. 
To this end, we leverage dense depth priors in order to constrain the NeRF optimization.
First, we take advantage of the sparse depth data that is freely available from the structure from motion (SfM) preprocessing step used to estimate camera poses.
Second, we use depth completion to convert these sparse points into dense depth maps and uncertainty estimates, which are used to guide NeRF optimization.
Our method enables data-efficient novel view synthesis on challenging indoor scenes, using as few as 18 images for an entire scene. \end{abstract}

\section{Introduction}
\label{sec:intro}
Synthesizing realistic views from varying viewpoints is essential for 
interactions between humans and virtual environments, hence it is of key importance for many 
virtual reality applications. The novel view synthesis task is especially relevant for indoor scenes, where it enables virtual navigation through buildings, tourist destinations, or game environments. 
When scaling up such applications, it is preferable to minimize the amount of input data required to store and process, as well as its acquisition time and cost. In addition, a static scene requirement is increasingly difficult to fulfill for a longer capture duration. Thus, our goal is novel view synthesis at room-scale using few input views. 

NeRF \cite{Mildenhall2020NeRFRS} represents the radiance field and density distribution of a scene as a multi-layer perceptron (MLP) and uses volume rendering to synthesize output images. This approach creates impressive, photo-realistic novel views of scenes with complex geometry and appearance. 
Unfortunately, applying NeRF to real-world, room-scale scenes given only tens of images does not produce desirable results (\cref{fig:teaser}) for the following reason: 
NeRF purely relies on RGB values to determine correspondences between input images. As a result, high visual quality is only achieved by NeRF when it is given enough images to overcome the inherent ambiguity of the correspondence problem. 
Real-world indoor scenes have characteristics that further complicate the ambiguity challenge: 
First, in contrast to an ``outside-in'' viewing scenario of images taken around a central object, views of rooms represent an ``inside-out'' viewing scenario, in which the same number of images will exhibit significantly less overlap with each other.
Second, indoor scenes often have large areas with minimal texture, such as white walls. 
Third, real-world data often has inconsistent color values across views, e.g., due to white balance or lens shading artifacts. 
These characteristics of indoor scenes are likewise challenging for SfM, leading to very sparse SfM reconstructions, often with severe outliers. 
Our idea is to use this noisy and incomplete depth data
and from it produce a complete dense map alongside a per-pixel uncertainty estimate of those depths, thereby increasing its value for NeRF --- especially in textureless, rarely observed, or color-inconsistent areas.

We propose a method that guides the NeRF optimization with dense depth priors, without the need for additional depth input (e.g., from an RGB-D sensor) of the scene. Instead, we take advantage of the sparse reconstruction that is freely available as a byproduct of running SfM to compute camera pose parameters. Specifically, we complete the sparse depth maps with a network that estimates uncertainty along with depth. Taking uncertainty into account, we use the resulting dense depth to constrain the optimization and to guide the scene sampling. We evaluate the effectiveness of our method on complete rooms from the Matterport3D \cite{Chang2017Matterport3DLF} and ScanNet \cite{Dai2017ScanNetR3} datasets, using only a handful of input images. We show that our approach improves over recent and concurrent work that uses sparse depth from SfM or multi-view stereo (MVS) in NeRF \cite{wei2021nerfingmvs,Deng2021DepthsupervisedNF}. 

In summary, we demonstrate that dense depth priors with uncertainty estimates enable novel view synthesis with NeRF on room-size scenes using only 18--36 images, enabled by the following contributions:
\begin{compactenum}
    \item A data-efficient approach to novel view synthesis on real-world scenes at room-scale. 
    \item An approach to enhance noisy sparse depth input from SfM to support the NeRF optimization.
    \item A technique for accounting for variable uncertainty when guiding NeRF with depth information.
\end{compactenum}

\section{Related Work}
\label{sec:related_work}
The ability to synthesize novel views of a scene from a set of observed images and corresponding camera viewpoints is necessary for enabling virtual experiences of real-world environments. 
In situations where it is feasible to densely sample images of the scene, 
novel viewpoints can be synthesized with simple light field interpolation~\cite{gortler96lumigraph,levoy96lightfield}. 
However, when fewer observed views of the scene are available, it becomes increasingly necessary to use information about the scene's geometry to render new views. A common paradigm for geometry-based view synthesis is to use a triangle mesh representation of scene geometry to reproject observed images into each novel viewpoint and combine them using either heuristic~\cite{buehler2001unstructured,debevec1992modeling,wood2000surface} or learned~\cite{hedman2019deep,riegler2020free} blending algorithms. More recently, these mesh-based geometry models have been replaced by volumetric scene representations such as voxel grids~\cite{lombardi2019neuralvolumes} or multiplane images~\cite{flynn2019deepview,mildenhall2019llff,srinivasan2019pushing,zhou2018stereomag}.
NeRF~\cite{Mildenhall2020NeRFRS} popularized an approach that avoids the steep scaling properties of discrete voxel representations by representing a scene as a continuous volume, parameterized by a MLP that is optimized to minimize the loss of re-rendering all observed views of a scene. Since its introduction, NeRF has become the dominant scene representation for view synthesis, and many recent works are built on top of NeRF's neural volumetric model. 

However, in situations where the scene is observed from very few sparsely-sampled viewpoints, NeRF's high capacity to model detailed geometry and appearance can 
result in various artifacts, such as ``floaters'', i.e., artifacts caused by a flawed density distribution. 
In this work, we directly address the few-input setting, proposing a strategy that takes advantage of sparse depth to constrain NeRF's scene geometry and improve rendering quality. This depth data is freely available as a byproduct of running SfM to compute camera poses from the input images (e.g., by using COLMAP~\cite{schoenberger2016sfm}).
Our method takes inspiration from techniques that generate complete dense depth maps from sparse depth inputs. These include classic techniques that fuse observed depths into a single 3D reconstruction, typically in the form of a truncated signed distance function~\cite{curlesslevoy,newcombe2016kinectfusion}, as well as more recent techniques that train deep networks to operate over the sparsely observed geometry in 3D~\cite{dai2018scancomplete,dai2020sgnn}.  Although these methods are effective for dense 3D scene reconstruction, the resulting geometry is not ideal for view synthesis since its edges frequently do not align with edges in the observed images. Instead, we leverage recent work on 2D sparse depth completion that directly completes depth maps in image space~\cite{cheng2018cspn,Cheng2020LearningDW} and extend it to also predict uncertainty. 

A few recent works have also proposed incorporating depth observations into NeRF reconstruction. NerfingMVS~\cite{wei2021nerfingmvs} uses depth from MVS to overfit a depth predictor to the scene. The resulting depth prior guides the NeRF sampling. In comparison, our method employs depth completion on the SfM depth and additionally employs a depth loss to supervise the geometry recovered by NeRF. This way, our novel views achieve significantly better color and depth quality in the few-input setting without relying on the computationally more expensive MVS preprocessing. Concurrent work on depth-supervised NeRF~\cite{Deng2021DepthsupervisedNF} directly uses sparse depth information from SfM in the NeRF optimization. To handle inaccuracy in the sparse reconstruction, the 3D points are weighted according to their reprojection error. In contrast, we learn dense depth priors with uncertainty to more effectively guide the optimization, leading to more detailed novel views, as well as more accurate geometry and higher robustness to SfM outliers.

\section{Method}
\label{sec:method}
\begin{figure*}[ht]
  \centering
  \includegraphics[width=\linewidth,trim={2.4cm 10.4cm 1.7cm 2.8cm},clip]{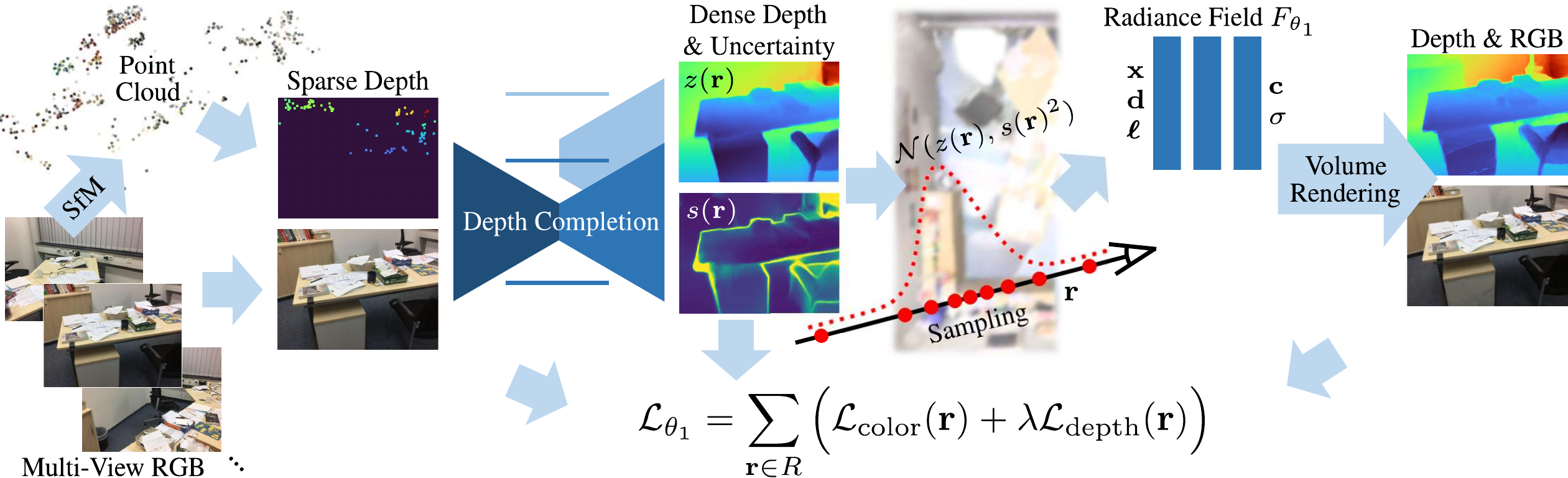}
   \caption{Overview of our radiance field optimization pipeline. Given a small set of RGB images of a room, we run SfM to obtain camera parameters and a sparse reconstruction, from which a sparse depth map is rendered for each input view. A depth completion network predicts dense depth and standard deviation, which is used to focus the scene sampling on surfaces. The samples on a ray, its viewing direction and a per-camera latent code are input to the radiance field. The output color and density are integrated to obtain the pixel's color and the ray's expected termination depth. The radiance field is supervised using the input RGB and the depth completion output. }
   \label{fig:pipeline}
\end{figure*}
Our method facilitates room-scale novel view synthesis from a small collection of RGB images $\{I_i\}^{N-1}_{i=0}$, $I_i \in [0, 1]^{H\times W\times3}$ (see \cref{fig:pipeline}). As a preprocessing step (e.g., using SfM), the camera pose $\mathbf{p}_i \in \mathbb{R}^6$, intrinsic parameters $K_i \in \mathbb{R}^{3\times3}$, and a sparse depth map $Z_i^{\mathrm{sparse}} \in [0, t_f]^{H\times W}$ are computed for each image. $0$ values of the sparse depth indicate invalid pixels, and $t_f$ is the far plane of the volume rendering. 
Our approach builds upon NeRF \cite{Mildenhall2020NeRFRS}. Prior to the NeRF optimization, a network estimates depth with uncertainty from the sparse depth input (\cref{ssec:depth_completion}). We incorporate the resulting dense depth prior into the NeRF optimization by adding a depth constraint and a guided sampling approach (\cref{ssec:depth_supervision}). 

\subsection{Depth Completion with Uncertainty}
\label{ssec:depth_completion}
\paragraph{Network Architecture} With the goal of completing sparse depth from SfM, two challenges presented by this input data play a key role in designing the depth prior network. First, sparse reconstructions are noisy and have outliers. As a consequence, dense depth predictions are expected to have spatially varying accuracy, which makes it crucial to know the uncertainty at a per-pixel level. 
Second, the density of SfM point clouds varies significantly across space, depending on the number of image features. 
E.g., SfM reconstructions from 18--20 images per ScanNet scene lead to sparse depth maps with 0.04\% valid pixels on average.
Hence, depth completion must be able to predict dense depth even from largely empty sparse depth maps.

In order to address the first challenge, we construct our depth prior network $D_{\theta_0}$ to predict dense depth maps $Z_i^{\mathrm{dense}} \in [0, t_f]^{H\times W}$ along with pixelwise standard deviations $S_i \in [0, \infty)^{H\times W}$ from the sparse depth maps: 
\begin{equation}
\left[Z_i^{\mathrm{dense}}, S_i\right] = D_{\theta_0}\!\left(I_i, Z_i^{\mathrm{sparse}}\right)\, ,
\label{eq:depth_completion}
\end{equation}
where $D_{\theta_0}$ is a convolutional network with ResNet~\cite{He2016DeepRL} downsampling and skip connections to two upsampling branches to predict depth $Z_i^{\mathrm{dense}}$ and standard deviation $S_i$. 
To address the second challenge of extremely sparse input depth, we employ a Convolutional Spatial Propagation Network (CSPN)~\cite{Cheng2020LearningDW} in each branch. This refinement block locally and iteratively applies a kernel with weights given by a learned affinity matrix. This refines the often blurry depth output to become more detailed and sharp. Equally important, this process spreads information to neighboring pixels; i.e., information propagates further with each iteration. An increased number of iterations in the depth and the uncertainty head prove helpful to handle very sparse input. 

\paragraph{Network Training} Though we evaluate on RGB-only data using SfM, we train our model on RGB-D data from ScanNet~\cite{Dai2017ScanNetR3} and Matterport3D~\cite{Chang2017Matterport3DLF}. 
To avoid both the effort of running SfM on a large dataset and the possibility of SfM failures in the training data, the sparse depth input is sampled from the range sensor depth.
As such, it is critical to subsample and perturb the dense depth from the sensor in a way that creates realistic sparse training depth to enable the network to generalize to real SfM input at test time. 
Specifically, a SIFT feature extractor, e.g., from COLMAP~\cite{schoenberger2016sfm}, is used to determine locations where sparse depth points would exist in a SfM reconstruction. We sample the sensor depth at these points and perturb it with Gaussian noise $\mathcal{N}(0, s_{\mathrm{noise}}(z)^2)$, where the standard deviation $s_{\mathrm{noise}}$ increases with depth. The function $s_{\mathrm{noise}}(z)$ is determined by fitting a second-order polynomial to the depth deviation between the sparse SfM reconstructions and sensor depth. 

Under the assumption that the output is normally distributed, we supervise the network by minimizing the negative log likelihood of a Gaussian: 
\begin{equation}
\mathcal{L}_{\theta_0} = \frac{1}{n} \sum_{j=1}^{n}\left(\log(s_j^2) + \frac{\left(z_j - z_{\mathrm{sensor},j}\right)^2}{s_j^2}\right),
\label{eq:depth_completion_loss}
\end{equation}
where $z_j$ and $s_j$ are the predicted depth and standard deviation of pixel $j$, $z_{\mathrm{sensor},j}$ is the sensor depth at $j$, and $n$ is the number of valid pixels in the dense sensor depth map. 

\subsection{Radiance Field with Dense Depth Priors}
\label{ssec:depth_supervision}
\paragraph{Scene Representation} Following NeRF \cite{Mildenhall2020NeRFRS}, we encode the radiance field of the scene into a MLP $F_{\theta_1}$ that predicts color $\mathbf{c} = [r, g, b]$ and volume density $\sigma$ from a position $\mathbf{x} \in \mathbb{R}^3$ and a unit-norm viewing direction $\mathbf{d} \in \mathbb{S}^2$. $\gamma$ applies a positional encoding with 9 frequencies on the position. Because our scenes are angularly undersampled, we minimize the capacity of our view-dependent network by omitting positional encoding for the viewing direction. 
\begin{equation}
\left[\mathbf{c}, \sigma\right] = F_{\theta_1}\!\left(\gamma(\mathbf{x}), \mathbf{d}, \boldsymbol{\ell}_i\right).
\label{eq:mlp}
\end{equation}
As an additional input to $F_{\theta_1}$, we generate a per-image embedding vector $\boldsymbol{\ell}_i \in \mathbb{R}^e$ \cite{MartinBrualla2021NeRFIT}. This allows the network to compensate for view-specific phenomena such as inconsistent lighting or lens shading, which can cause severe artifacts in novel views, particularly with few input images. 

\paragraph{Optimization with Depth Constraint} To optimize the radiance field, the color $\hat{\mathbf{C}}(\mathbf{r})$ of each pixel in the batch $R$ is computed by evaluating a discretized version of the volume rendering integral (\cref{eq:volume_rendering} \cite{Mildenhall2020NeRFRS}). 
Specifically, a pixel determines a ray $\mathbf{r}(t) = \mathbf{o} + t\mathbf{d}$ whose origin is at the camera's center of projection $\mathbf{o}$. Rays are sampled along their traversal through the volume. For each sampling location $t_k \in [t_n, t_f]$ within the near and far planes, a query to $F_{\theta_1}$ provides the local color and density. 
\begin{align}
\hat{\mathbf{C}}(\mathbf{r}) &= \sum_{k=1}^{K} w_k\mathbf{c}_k \, , \label{eq:volume_rendering} \\
\text{where} \quad w_k &= T_k\left(1 - \exp(-\sigma_k\delta_k)\right) \, ,
\label{eq:rendering_weight} \\
T_k &= \exp \left( -\sum_{k'=1}^{k} \sigma_{k'}\delta_{k'}\right)\, , \\ 
\delta_k &= t_{k+1} - t_k \, .
\end{align}
Besides the predicted color of a ray, a NeRF depth estimate $\hat{z}(\mathbf{r})$ and standard deviation $\hat{s}(\mathbf{r})$ are needed to supervise the radiance field according to the learned depth prior (\cref{ssec:depth_completion}). The NeRF depth estimate and standard deviation are computed from the rendering weights $w_k$: 
\begin{align}
    \hat{z}(\mathbf{r}) &= \sum_{k=1}^{K}w_kt_k \, , \quad
    \hat{s}(\mathbf{r})^2 = \sum_{k=1}^{K}w_k(t_k - \hat{z}(\mathbf{r}))^2.
\end{align}
The network parameters $\theta_1$ are optimized using a loss function $\mathcal{L}_{\theta_1}$ composed of a mean squared error (MSE) term on the color output $\mathcal{L}_{\mathrm{color}}$ and a Gaussian negative log likelihood (GNLL) term on the depth output $\mathcal{L}_{\mathrm{depth}}$: 
\begin{align}
    \mathcal{L}_{\theta_1} &= \displaystyle\sum_{\mathbf{r}\in R}\Big(\mathcal{L}_{\mathrm{color}}(\mathbf{r}) + \lambda \mathcal{L}_{\mathrm{depth}}(\mathbf{r})\Big), \\
    \mathcal{L}_{\mathrm{color}}(\mathbf{r}) &= \begin{Vmatrix} \hat{\mathbf{C}}(\mathbf{r}) - \mathbf{C}(\mathbf{r}) \end{Vmatrix}_2^2,
\end{align}
\begin{align}
    \!\!\!\mathcal{L}_{\mathrm{depth}}(\mathbf{r}) = &\begin{cases} \log\left(\hat{s}(\mathbf{r})^2\right) + \frac{\left(\hat{z}(\mathbf{r}) - z(\mathbf{r})\right)^2}{\hat{s}(\mathbf{r})^2} &\!\! \text{if } P \text{ or } Q \\
    \label{eq:depth_loss}
    0 &\!\! \text{otherwise,} \end{cases} \\
    \!\!\!\!\!\text{where} \quad P &= | \hat{z}(\mathbf{r}) - z(\mathbf{r}) | > s(\mathbf{r}) \, , \label{eq:conditionA}\\
    \!\!\!\!\!Q &= \hat{s}(\mathbf{r}) > s(\mathbf{r}) \, . \label{eq:conditionB}
\end{align}
Here $z(\mathbf{r})$ and $s(\mathbf{r})$ are the target depth and standard deviation from the corresponding $Z_i^{\mathrm{dense}}$ and $S_i$. 
The depth loss is applied to rays where at least one of the following conditions is true: 
1) the difference between the predicted and target depth is greater than the target standard deviation \cref{eq:conditionA}, or
2) the predicted standard deviation is greater than the target standard deviation \cref{eq:conditionB}. 
This way, the loss encourages NeRF to terminate rays within one standard deviation of the most certain surface observation in the depth prior. 
At the same time, NeRF retains some freedom to allocate density to best minimize the color loss. The effectiveness of this depth loss in contrast to MSE is shown in the ablation study (\cref{ssec:ablation_studies}). 

\paragraph{Depth-Guided Sampling} In addition to the depth loss function, the depth prior contains valuable signal to guide sampling along a ray. To render one pixel of a room-scale scene, we require the same number of MLP queries as the original NeRF; however, we replace the coarse network used for hierarchical sampling. During optimization, half of the samples are distributed between the near and far planes and the second half are drawn from the Gaussian distribution determined by the depth prior $\mathcal{N}(z(\mathbf{r}), s(\mathbf{r})^2)$. 
At test time, when the depth is unknown, the first half of the samples are used to render an approximate depth $\hat{z}(\mathbf{r})$ and standard deviation $\hat{s}(\mathbf{r})$ that is then used to sample the second half according to $\mathcal{N}(\hat{z}(\mathbf{r}), \hat{s}(\mathbf{r})^2)$. 

\section{Results}
\label{sec:experiments}
We evaluate our method  with a baseline comparison (\cref{ssec:baseline_comparison}) and an ablation study (\cref{ssec:ablation_studies}) on the ScanNet~\cite{Dai2017ScanNetR3} and Matterport3D~\cite{Chang2017Matterport3DLF} datasets. 
\subsection{Experimental Setup}
\paragraph{ScanNet} 
We run COLMAP SfM~\cite{schoenberger2016sfm} to obtain camera parameters and sparse depth. Specifically, we run SfM on all images to obtain camera parameters. To ensure a clean split between train and test data, we withhold the test images when computing the point cloud used for rendering the sparse depth maps. On average, the resulting depth maps have 0.04\% valid pixels. 
We use three sample scenes each with 18 to 20 train images and 8 test images. 
This set of images results from excluding video frames with motion blur 
while ensuring that surfaces are observed from at least one input view. Details are provided in \cref{ssec:datasets_scannet}. 

\paragraph{Matterport3D}
Using RGB images from the PrimeSense camera, COLMAP SfM struggled to reconstruct complete rooms in Matterport3D, hence, we mimic sparse depth from SfM by sampling and perturbing the sensor depth as described for depth prior training in \cref{ssec:depth_completion}. Sparse depth maps rendered from a SfM point cloud are by nature 3D-consistent. While consistency in 3D is irrelevant for training 2D depth completion, it plays a critical role when optimizing a 3D scene representation with NeRF. Hence, we ensure 3D-consistent sparse depth on the scenes used for NeRF by projecting the sampled and perturbed 3D points to all other views. On average the resulting depth maps are 0.1\% complete. 
The impact of the sparse depth density is studied in \cref{sec:sparse_depth_density}. 
We evaluate three example rooms each with 24 to 36 train images and 8 test images. 

\paragraph{NeRF Optimization}
We process rays in batches of 1024 and use the Adam optimizer~\cite{Kingma2015AdamAM} with learning rate 0.0005. 
For fairness, all approaches in the ablation and baseline experiments are configured to use 256 MLP evaluations per pixel, independent of the used sampling approach. The radiance fields are optimized for 500k iterations.  
Further NeRF and depth prior implementation details are available in \cref{sec:implementation_details}. 

\paragraph{Evaluation Metrics}
For quantitative comparison, we compute the peak signal-to-noise ratio (PSNR), the Structural Similarity Index Measure (SSIM) \cite{Wang2004ImageQA} and the Learned Perceptual Image Patch Similarity (LPIPS) \cite{Zhang2018TheUE} on the RGB of novel views as well as the root-mean-square error (RMSE) on the expected ray termination depth of NeRF against the sensor depth in meters. 
By comparing color values directly, PSNR has limited expressiveness, when the images of the scene have inconsistent color. 
As shown in \cref{ssec:ablation_studies}, the latent codes used to represent view-specific appearance largely help to produce consistent colors across the scene. Still, the color of a rendered image will not necessarily be similar to that of the test view against which it is evaluated. To compensate for this difference, we report an additional PSNR value, which is computed after optimizing for the latent codes on the entire test views. We are unable to use the left/right image split evaluation procedure from NeRF-W \cite{MartinBrualla2021NeRFIT}, because appearance changes too drastically across the image, so these numbers should be considered an upper bound on performance. 
This additional metric is listed in parentheses (\cref{tab:results_scannet,tab:results_matterport}) for all approaches that use a latent code. 
All other metrics as well as all renderings in the paper are computed by setting the latent code to zero, given that the codes are unknown at test time. 
\subsection{Depth Priors}
\cref{tab:depth_priors} shows the depth prior accuracy on the three ScanNet and three Matterport3D scenes used for NeRF. These scenes are part of the test sets during depth completion training. 
\begin{table}[tb]
  \centering
  \small
  \begin{tabular}{@{}lcc@{}}
    \toprule
    & \multicolumn{2}{c}{RMSE [m] $\downarrow$} \\
    Dataset & Sparse depth  & Dense depth \\
    \midrule
    ScanNet & 0.261 & 0.268 \\
    Matterport3D & 0.041 & 0.135 \\
    \bottomrule
  \end{tabular}
  \caption{Accuracy of depth priors.}
  \vspace{-.3cm}
  \label{tab:depth_priors}
\end{table}
The higher quality, generated sparse depth on Matterport3D leads to more accurate dense depth priors. However, the network interpolates the more noisy sparse depth from SfM on ScanNet without a relevant drop in accuracy. 
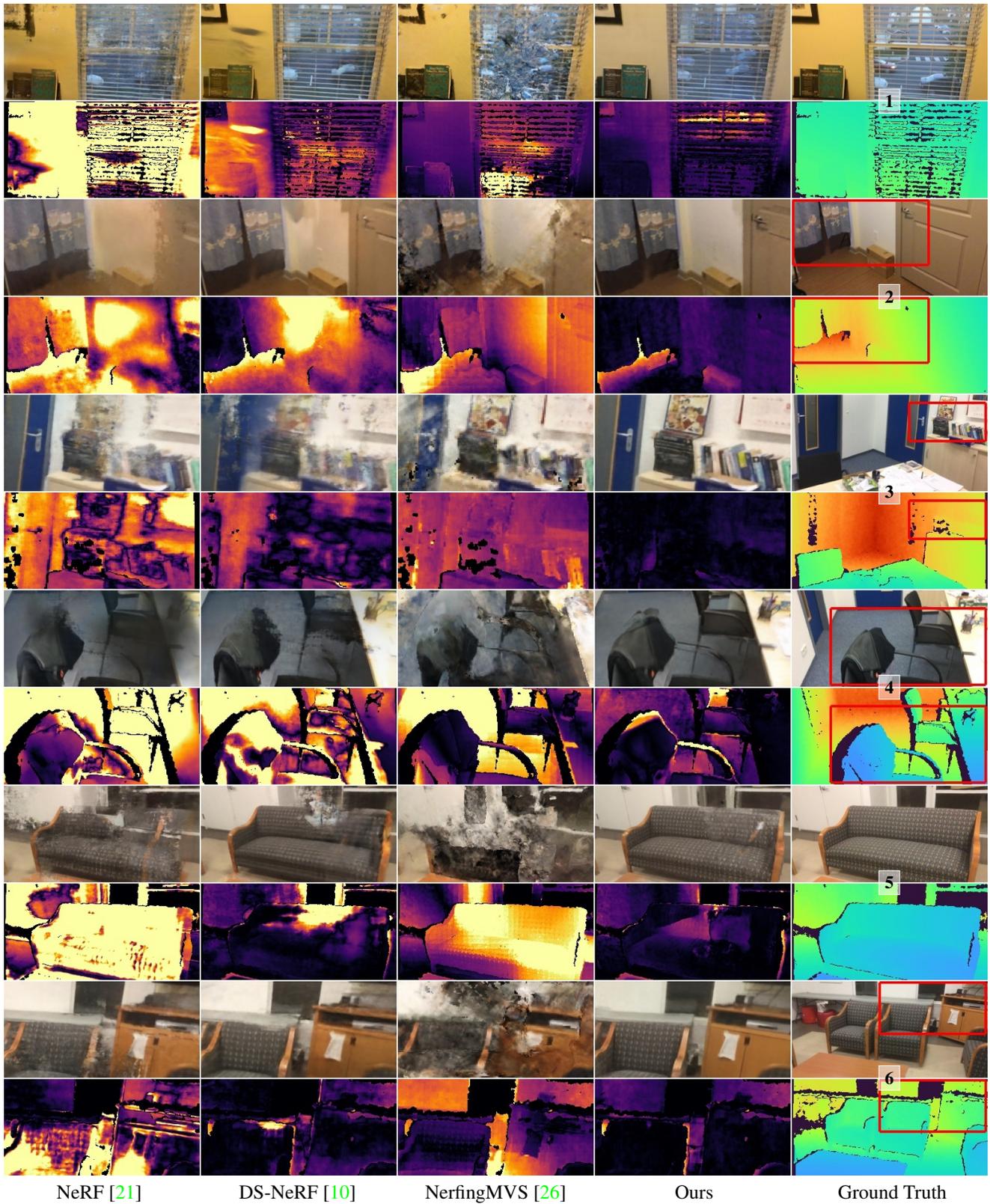
\begin{figure*}[htbp]
\centering
\begin{tabular}{@{\,\,\,\,}p{0.195\linewidth}@{\,\,}p{0.195\linewidth}@{\,\,}p{0.195\linewidth}@{\,\,}p{0.195\linewidth}@{\,\,}p{0.195\linewidth}@{\,\,\,\,}}
    \multicolumn{5}{c}{
    \begin{tikzpicture}[squarednode/.style={rectangle, draw=white, fill=white, very thin, minimum size=2mm, text opacity=1,fill opacity=0.5}]
        \node[anchor=south west,inner sep=0] (image) at (0,0)
        {\includegraphics[width=\linewidth]{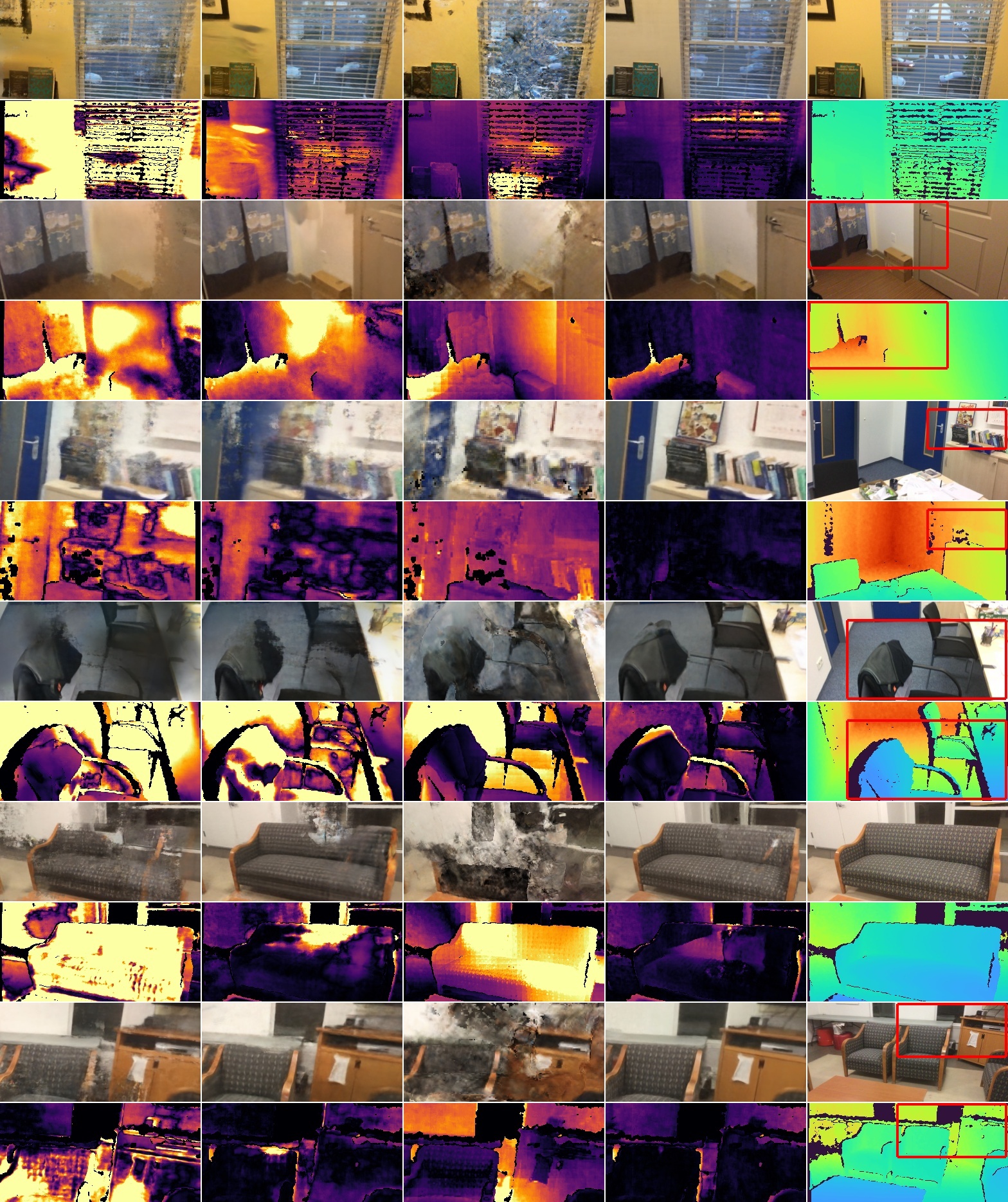}};
        \begin{scope}[x=(image.south east),y=(image.north west)]
        \bf
        \small
        \node[squarednode] at (0.9, 11/12) (a) {1};
        \node[squarednode] at (0.9, 9/12) (b) {2};
        \node[squarednode] at (0.9, 7/12) (c) {3};
        \node[squarednode] at (0.9, 5/12) (d) {4};
        \node[squarednode] at (0.9, 3/12) (e) {5};
        \node[squarednode] at (0.9, 1/12) (f) {6};
        \end{scope}
    \end{tikzpicture}
    } \\
    \centering{NeRF \cite{Mildenhall2020NeRFRS}} & \centering{DS-NeRF \cite{Deng2021DepthsupervisedNF}} & \centering{NerfingMVS \cite{wei2021nerfingmvs}} & \centering{Ours} & \centering{Ground Truth} \\
\end{tabular}
\caption{Rendered RGB and depth error for test views from three ScanNet rooms next to the ground truth RGB and depth.}
\label{fig:results_scannet}
\end{figure*}
\begin{figure*}[b]
\centering
\begin{tabular}{@{\,\,\,\,}p{0.136\linewidth}@{\,\,}p{0.136\linewidth}@{\,\,}p{0.136\linewidth}@{\,\,}p{0.136\linewidth}@{\,\,}p{0.136\linewidth}@{\,\,}p{0.136\linewidth}@{\,\,}p{0.136\linewidth}@{\,\,\,\,}}
    \multicolumn{7}{c}{
    \begin{tikzpicture}[squarednode/.style={rectangle, draw=white, fill=white, very thin, minimum size=2mm, text opacity=1,fill opacity=0.5}]
        \node[anchor=south west,inner sep=0] (image) at (0,0)
        {\includegraphics[width=\linewidth]{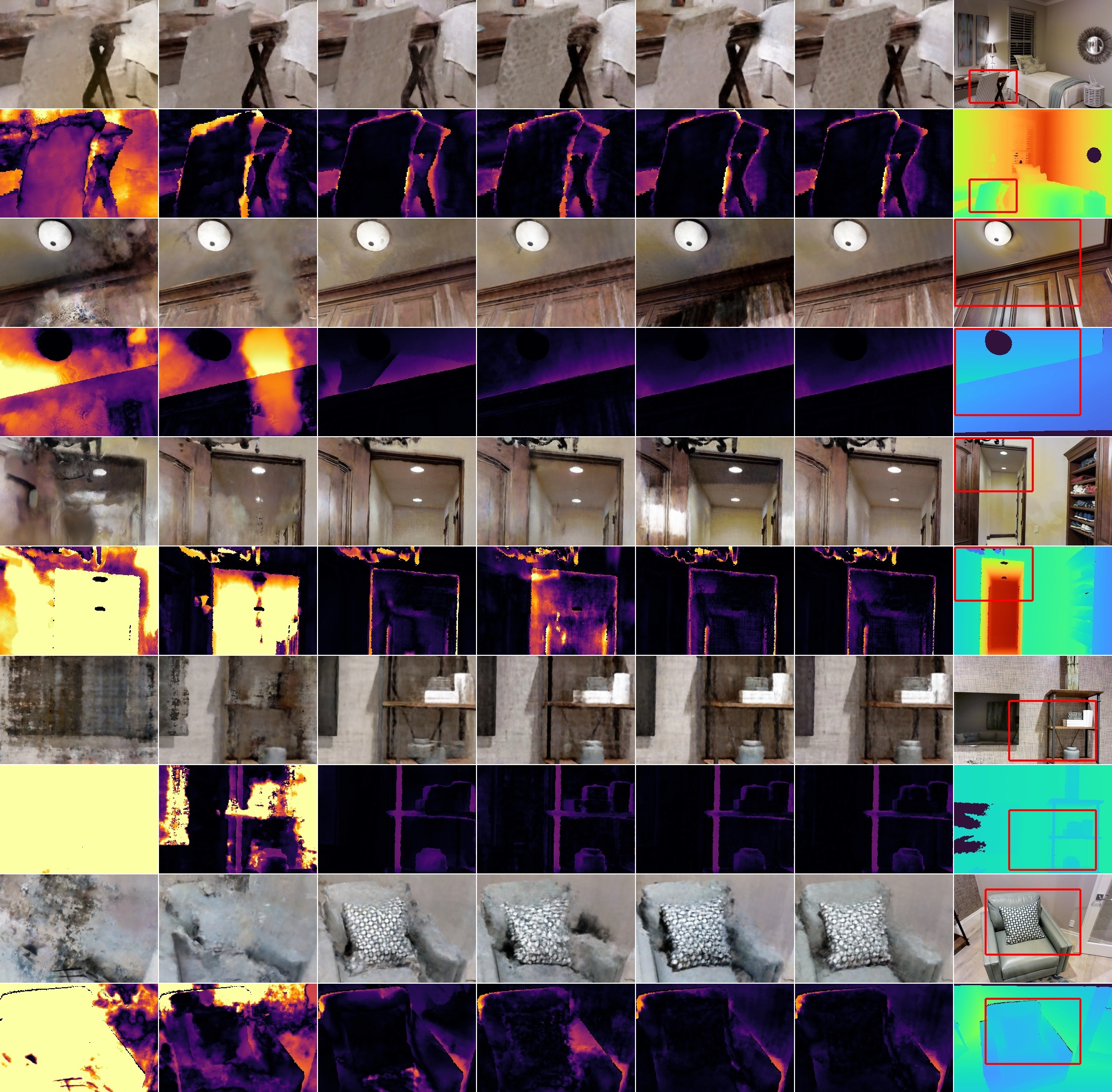}};
        \begin{scope}[x=(image.south east),y=(image.north west)]
        \bf
        \small
        \node[squarednode] at (13/14, 9/10) (a) {1};
        \node[squarednode] at (13/14, 7/10) (b) {2};
        \node[squarednode] at (13/14, 5/10) (c) {3};
        \node[squarednode] at (13/14, 3/10) (d) {4};
        \node[squarednode] at (13/14, 1/10) (e) {5};
        \end{scope}
    \end{tikzpicture}
    } \\
    \centering{NeRF \cite{Mildenhall2020NeRFRS}} & \centering{Ours w/o Completion} & \centering{Ours w/o Uncertainty} & \centering{Ours w/o GNLL} & \centering{Ours w/o Latent Code} & \centering{Ours} & \centering{Ground Truth} \\
\end{tabular}
\caption{Rendered RGB and depth error for test views from three Matterport3D rooms next to the ground truth RGB and depth.}
\label{fig:results_matterport}
\end{figure*}
\subsection{Baseline Comparison}
\label{ssec:baseline_comparison}
We compare our method to NeRF \cite{Mildenhall2020NeRFRS} as well as recent and concurrent work that equally uses sparse depth input in NeRF, namely Depth-supervised NeRF (DS-NeRF) \cite{Deng2021DepthsupervisedNF} and NerfingMVS \cite{wei2021nerfingmvs}. Since DS-NeRF and NerfingMVS rely on SfM and MVS depth, respectively, they are run on ScanNet. NeRF and our method are run on both datasets. 
The quantitative results (\cref{tab:results_scannet}) show that our method outperforms the baselines in all metrics. 

``Floaters'' are a common problem when applying NeRF approaches in a setting with few input views. 
By using dense depth priors with uncertainty, our method strongly reduces these artifacts compared to the baselines (example 2 \cref{fig:results_scannet}). 
This contributes to far more accurate depth output and greater detail in color, e.g., visible in the books and the door handle (example 3 \cref{fig:results_scannet}). We found that our method is more robust to outliers in the sparse depth input. E.g., erroneous SfM points  in the area of the sofa back (example 5 \cref{fig:results_scannet}) cause much larger deficiencies in geometry and color of the other approaches. This suggests that dense depth priors with uncertainty focus the optimization on more certain and accurate views, while direct incorporation of sparse depth, as in DS-NeRF, is more error-prone. 
Besides greater robustness to outliers, dense depth guides NeRF better at object boundaries that are not represented in the sparse depth input. This is observable in example 6 (\cref{fig:results_scannet}), where a part of the chair back is missing in DS-NeRF, while it is complete using our method. 

\paragraphNoSpace{NerfingMVS Details}
The error map calculation in NerfingMVS fails when applied to an entire room as opposed to a local region, causing invalid sampling ranges. The issue is solved as detailed in \cref{sec:implementation_details_nerfingmvs}. 
To improve the performance of this baseline, we train its depth predictor 10 epochs longer than was done in the paper. 
Still, the depth priors on the ScanNet scenes remain at RMSE 0.379m. While our method uses only train images to compute the sparse depth input, NerfingMVS runs COLMAP MVS on train and test images together, which gives them an advantage.

\begin{table}[t]
  \centering
\resizebox{\linewidth}{!}{
  \begin{tabular}{@{}lc@{\,\,}c@{\,\,}c@{\,\,}c@{\,\,}c@{}}
    \toprule
    & & & & & Depth \\
    Method & \multicolumn{2}{c}{PSNR$\uparrow$} & SSIM$\uparrow$ & LPIPS$\downarrow$ & RMSE $\downarrow$ \\
    \midrule
    NeRF \cite{Mildenhall2020NeRFRS} & 19.03 & & 0.670 & 0.398 & 1.163 \\
    DS-NeRF \cite{Deng2021DepthsupervisedNF} & 20.85 & & 0.713 & 0.344 & 0.447 \\
    NerfingMVS \cite{wei2021nerfingmvs} & 16.29 & & 0.626 & 0.502 & 0.482 \\
    Ours w/o Completion & 20.43 & (22.10) & 0.707 & 0.366 & 0.526 \\
    Ours w/o Uncertainty & 20.09 & (22.21) & 0.714 & 0.308 & 0.279 \\
    Ours w/o GNLL & 20.80 & (22.23) & 0.733 & 0.312 & 0.275 \\
    Ours w/o Latent Code & 20.87 & & 0.726 & \textbf{0.293} & 0.243 \\
    Ours & \textbf{20.96} & (\textbf{22.30}) & \textbf{0.737} & 0.294 & \textbf{0.236} \\
    \bottomrule
  \end{tabular}
  }
  \caption{Quantitative results on ScanNet.
  Parentheses indicate PSNRs obtained after optimizing a latent code, when applicable.}
  \vspace{-.3cm}
  \label{tab:results_scannet}
\end{table}

\subsection{Ablation Study}
\label{ssec:ablation_studies}
To verify the effectiveness of the added components, we conduct ablation experiments on the ScanNet and Matterport3D scenes. 
The quantitative results (\cref{tab:results_scannet,tab:results_matterport}) show that the full version of our method achieves the best performance in image quality and depth estimates. 
This is consistent with the qualitative results in \cref{fig:results_matterport}. 

\paragraphNoSpace{Without Completion}
Omitting depth completion and supervising with sparse depth only leads to inaccurate depth and color due to ``floaters'' in areas without depth input. 
Even in areas with sparse depth points, the results are less sharp than in versions that use completed depth. 

\paragraphNoSpace{Without Uncertainty}
Removing uncertainty from the optimization causes problems in resolving inconsistency in overlapping areas of the 2D depth priors. This results in wrong edges in RGB and depth (example 2 \cref{fig:results_matterport}), duplication artifacts (example 4 \cref{fig:results_matterport}) or lacking detail, e.g., in the patterns on the back of the chair (example 1 \cref{fig:results_matterport}). 
The quantitative results on ScanNet (\cref{tab:results_scannet}) show that considering uncertainty becomes even more important when using the lower quality sparse depth from SfM. 

\paragraphNoSpace{Without GNLL}
In this experiment, we replace GNLL with MSE in our depth loss (\cref{eq:depth_loss}), and observe that MSE struggles to constrain density behind surfaces. The lack of sharp edges in the density distribution is most visible for novel views looking in tangential direction of a surface, e.g., looking into the corridor (example 3 \cref{fig:results_matterport}). 

\paragraphNoSpace{Without Latent Code}
Omitting the latent code that models per-camera information, leads to incapability to produce smooth and consistent color output across the scene. When rendering novel views, the frustums of training images are clearly visible by causing severe shifts in color intensity (examples 2 and 3, \cref{fig:results_matterport}). 

\begin{table}[tb]
  \centering
\resizebox{\linewidth}{!}{
  \begin{tabular}{@{}lc@{\,\,}c@{\,\,}c@{\,\,}c@{\,\,}c@{}}
    \toprule
    & & & & & Depth \\
    Method & \multicolumn{2}{c}{PSNR$\uparrow$} & SSIM$\uparrow$ & LPIPS$\downarrow$ & RMSE $\downarrow$ \\
    \midrule
    NeRF \cite{Mildenhall2020NeRFRS} & 15.24 & & 0.531 & 0.610 & 1.362 \\
    Ours w/o Completion & 16.90 & (18.84) & 0.615 & 0.521 & 0.427 \\
    Ours w/o Uncertainty & 17.95 & (20.37) & 0.658 & 0.413 & 0.115 \\
    Ours w/o GNLL & 18.00 & (20.65) & 0.669 & 0.423 & 0.133 \\
    Ours w/o Latent Code & 17.42 & & 0.654 & 0.410 & \textbf{0.110} \\
    Ours & \textbf{18.33} & (\textbf{20.82}) & \textbf{0.673} & \textbf{0.402} & 0.114 \\
    \bottomrule
  \end{tabular}
  }
  \caption{Quantitative results on Matterport3D, using the same format as \cref{tab:results_scannet}.}
  \vspace{-.3cm}
  \label{tab:results_matterport}
\end{table}

\subsection{Limitations}
Our method allows for a significant reduction in the number of input images for NeRF-based novel view synthesis while at the same time applying it to larger room-size scenes. However, other NeRF limitations 
such as long optimization times and slow rendering remain. 
As a consequence of 
the drastic reduction in the number of input images, surfaces are typically not observed by more than two other views, hence view-dependent effects are limited. 
While our approach optimizes NeRF given as few as 18 images, the depth prior network requires a larger training dataset.
Although these priors generalize well and only need to be trained once, it would be beneficial if the depth reconstruction could be also learned from a sparse setting.

\section{Conclusion}
We have presented a method for novel view synthesis using neural radiance fields (NeRF) that leverages dense depth priors, thus facilitating reconstructions with only 18 to 36 input images for a complete room. 
By learning a depth prior that generalizes across scenes, our method takes advantage of depth information without requiring depth sensor input of the scene. Instead, the depth prior network relies on the sparse reconstruction, which is available for free after structure from motion (SfM) on the input images. With only a few input views available, we show that our dense depth priors with uncertainty effectively guide the NeRF optimization, thus leading to significantly higher image quality of novel views and more accurate depth estimates compared to other approaches using SfM or multi-view stereo output in NeRF. Overall, we believe that our method is an important step towards making NeRF reconstructions available in commodity settings.

\section*{Acknowledgements}
This project is funded by a TUM-IAS Rudolf Mößbauer Fellowship, the ERC Starting Grant Scan2CAD (804724), and the German Research Foundation (DFG) Grant Making Machine Learning on Static and Dynamic 3D Data Practical.
We thank Angela Dai for the video voice-over.

{\small
\bibliographystyle{ieee_fullname}
\bibliography{references}

\begin{thebibliography}{10}\itemsep=-1pt

\bibitem{buehler2001unstructured}
Chris Buehler, Michael Bosse, Leonard McMillan, Steven Gortler, and Michael
  Cohen.
\newblock Unstructured lumigraph rendering.
\newblock In {\em SIGGRAPH}, pages 425--432, 2001.

\bibitem{Chang2017Matterport3DLF}
Angel~X. Chang, Angela Dai, Thomas~A. Funkhouser, Maciej Halber, Matthias
  Nie{\ss}ner, Manolis Savva, Shuran Song, Andy Zeng, and Yinda Zhang.
\newblock Matterport3d: Learning from rgb-d data in indoor environments.
\newblock {\em 3DV}, 2017.

\bibitem{cheng2018cspn}
Xinjing Cheng, Peng Wang, and Ruigang Yang.
\newblock Depth estimation via affinity learned with convolutional spatial
  propagation network.
\newblock In {\em Proceedings of the European Conference on Computer Vision
  (ECCV)}, pages 103--119, 2018.

\bibitem{Cheng2020LearningDW}
Xinjing Cheng, Peng Wang, and Ruigang Yang.
\newblock Learning depth with convolutional spatial propagation network.
\newblock {\em IEEE TPAMI}, 42, 2020.

\bibitem{curlesslevoy}
Brian Curless and Marc Levoy.
\newblock A volumetric method for building complex models from range images.
\newblock {\em Proceedings of the 23rd annual conference on Computer graphics
  and interactive techniques}, 1996.

\bibitem{Dai2017ScanNetR3}
Angela Dai, Angel~X. Chang, Manolis Savva, Maciej Halber, Thomas~A. Funkhouser,
  and Matthias Nie{\ss}ner.
\newblock Scannet: Richly-annotated 3d reconstructions of indoor scenes.
\newblock {\em CVPR}, 2017.

\bibitem{dai2020sgnn}
Angela Dai, Christian Diller, and Matthias Niessner.
\newblock Sg-nn: Sparse generative neural networks for self-supervised scene
  completion of rgb-d scans.
\newblock In {\em Proceedings of the IEEE/CVF Conference on Computer Vision and
  Pattern Recognition (CVPR)}, June 2020.

\bibitem{dai2018scancomplete}
Angela Dai, Daniel Ritchie, Martin Bokeloh, Scott Reed, J{\"u}rgen Sturm, and
  Matthias Nie{\ss}ner.
\newblock Scancomplete: Large-scale scene completion and semantic segmentation
  for 3d scans.
\newblock In {\em Proc. Computer Vision and Pattern Recognition (CVPR), IEEE},
  2018.

\bibitem{debevec1992modeling}
Paul~E. Debevec, Camillo~J. Taylor, and Jitendra Malik.
\newblock Modeling and rendering architecture from photographs: A hybrid
  geometry- and image-based approach.
\newblock In {\em Proceedings of the 23rd Annual Conference on Computer
  Graphics and Interactive Techniques}, SIGGRAPH '96, page 11–20, 1996.

\bibitem{Deng2021DepthsupervisedNF}
Kangle Deng, Andrew Liu, Jun-Yan Zhu, and Deva Ramanan.
\newblock Depth-supervised nerf: Fewer views and faster training for free.
\newblock {\em ArXiv}, abs/2107.02791, 2021.

\bibitem{flynn2019deepview}
John Flynn, Michael Broxton, Paul Debevec, Matthew DuVall, Graham Fyffe, Ryan
  Overbeck, Noah Snavely, and Richard Tucker.
\newblock Deepview: View synthesis with learned gradient descent.
\newblock In {\em Proc. Proceedings of the IEEE/CVF Conference on Computer
  Vision and Pattern Recognition}, pages 2367--2376, 2019.

\bibitem{gortler96lumigraph}
Steven~J Gortler, Radek Grzeszczuk, Richard Szeliski, and Michael~F Cohen.
\newblock The lumigraph.
\newblock In {\em SIGGRAPH}, pages 43--54, 1996.

\bibitem{He2016DeepRL}
Kaiming He, X. Zhang, Shaoqing Ren, and Jian Sun.
\newblock Deep residual learning for image recognition.
\newblock {\em CVPR}, 2016.

\bibitem{hedman2019deep}
Peter Hedman, Julien Philip, True Price, Jan-Michael Frahm, George Drettakis,
  and Gabriel Brostow.
\newblock Deep blending for free-viewpoint image-based rendering.
\newblock 2018.

\bibitem{newcombe2016kinectfusion}
Shahram Izadi, David Kim, Otmar Hilliges, David Molyneaux, Richard Newcombe,
  Pushmeet Kohli, Jamie Shotton, Steve Hodges, Dustin Freeman, Andrew Davison,
  and Andrew Fitzgibbon.
\newblock Kinectfusion: Real-time 3d reconstruction and interaction using a
  moving depth camera.
\newblock In {\em UIST '11 Proceedings of the 24th annual ACM symposium on User
  interface software and technology}, pages 559--568. ACM, October 2011.

\bibitem{Kingma2015AdamAM}
Diederik~P. Kingma and Jimmy Ba.
\newblock Adam: A method for stochastic optimization.
\newblock {\em CoRR}, abs/1412.6980, 2015.

\bibitem{levoy96lightfield}
Marc Levoy and Pat Hanrahan.
\newblock Light field rendering.
\newblock In {\em SIGGRAPH}, 1996.

\bibitem{lombardi2019neuralvolumes}
Stephen Lombardi, Tomas Simon, Jason Saragih, Gabriel Schwartz, Andreas
  Lehrmann, and Yaser Sheikh.
\newblock Neural volumes: Learning dynamic renderable volumes from images.
\newblock {\em ACM Transactions on Graphics (SIGGRAPH)}, 2019.

\bibitem{MartinBrualla2021NeRFIT}
Ricardo Martin-Brualla, Noha Radwan, Mehdi S.~M. Sajjadi, Jonathan~T. Barron,
  Alexey Dosovitskiy, and Daniel Duckworth.
\newblock Nerf in the wild: Neural radiance fields for unconstrained photo
  collections.
\newblock In {\em CVPR}, 2021.

\bibitem{mildenhall2019llff}
Ben Mildenhall, Pratul~P. Srinivasan, Rodrigo Ortiz-Cayon, Nima~K. Kalantari,
  Ravi Ramamoorthi, Ren Ng, and Abhishek Kar.
\newblock Local light field fusion: Practical view synthesis with prescriptive
  sampling guidelines.
\newblock {\em ACM Transactions on Graphics (SIGGRAPH)}, 2019.

\bibitem{Mildenhall2020NeRFRS}
Ben Mildenhall, Pratul~P. Srinivasan, Matthew Tancik, Jonathan~T. Barron, Ravi
  Ramamoorthi, and Ren Ng.
\newblock Nerf: Representing scenes as neural radiance fields for view
  synthesis.
\newblock In {\em ECCV}, 2020.

\bibitem{riegler2020free}
Gernot Riegler and Vladlen Koltun.
\newblock Free view synthesis.
\newblock In {\em ECCV}, 2020.

\bibitem{schoenberger2016sfm}
Johannes~Lutz Sch\"{o}nberger and Jan-Michael Frahm.
\newblock Structure-from-motion revisited.
\newblock In {\em Conference on Computer Vision and Pattern Recognition
  (CVPR)}, 2016.

\bibitem{srinivasan2019pushing}
Pratul~P. Srinivasan, Richard Tucker, Jonathan~T. Barron, Ravi Ramamoorthi, Ren
  Ng, and Noah Snavely.
\newblock Pushing the boundaries of view extrapolation with multiplane images.
\newblock In {\em CVPR}, 2019.

\bibitem{Wang2004ImageQA}
Zhou Wang, Alan~Conrad Bovik, Hamid~R. Sheikh, and Eero~P. Simoncelli.
\newblock Image quality assessment: from error visibility to structural
  similarity.
\newblock {\em IEEE Transactions on Image Processing}, 13:600--612, 2004.

\bibitem{wei2021nerfingmvs}
Yi Wei, Shaohui Liu, Yongming Rao, Wang Zhao, Jiwen Lu, and Jie Zhou.
\newblock Nerfingmvs: Guided optimization of neural radiance fields for indoor
  multi-view stereo.
\newblock In {\em ICCV}, 2021.

\bibitem{wood2000surface}
Daniel~N Wood, Daniel~I Azuma, Ken Aldinger, Brian Curless, Tom Duchamp,
  David~H Salesin, and Werner Stuetzle.
\newblock Surface light fields for 3d photography.
\newblock In {\em Proceedings of the 27th annual conference on Computer
  graphics and interactive techniques}, pages 287--296, 2000.

\bibitem{Zhang2018TheUE}
Richard Zhang, Phillip Isola, Alexei~A. Efros, Eli Shechtman, and Oliver Wang.
\newblock The unreasonable effectiveness of deep features as a perceptual
  metric.
\newblock {\em 2018 IEEE/CVF Conference on Computer Vision and Pattern
  Recognition}, pages 586--595, 2018.

\bibitem{zhou2018stereomag}
Tinghui Zhou, Richard Tucker, John Flynn, Graham Fyffe, and Noah Snavely.
\newblock Stereo magnification: Learning view synthesis using multiplane
  images.
\newblock {\em ACM Transactions on Graphics (SIGGRAPH)}, 2018.

\end{thebibliography}
}
\clearpage
\appendix
\section{Datasets}
\subsection{ScanNet \cite{Dai2017ScanNetR3}}
\label{ssec:datasets_scannet}
\paragraphNoSpace{Motion Blur Detection}
We consider motion blur when sampling a small subset of images to be used in NeRF: From each window of $n$ consecutive video frames the sharpest one is selected according to the following metric, where high values indicate sharpness: first, an image is converted to grayscale, then it is convolved with a discrete Laplacian kernel; finally, the variance is computed. 
$n$ is set to 10 or 20, depending on how densely the video samples the scene. 

\paragraphNoSpace{Train/Test Image Selection}
After removing images with severe motion blur, we consider the following criteria: 1) SfM successfully registers the set of images. 2) Surfaces to be reconstructed are observed from at least one input view. 
In practice, images are removed if their content is visible by other images and the remaining set fulfils 1). 
This way, 22\% of the train pixels are not observed by any other train view, 31\% are observed by one other, 47\% by two or more. Test views have on average 66\% overlap with their most overlapping train view. 

\paragraphNoSpace{Image Resolution}
The image resolution is 468$\times$624 after downsampling and cropping dark borders from calibration. 

\paragraphNoSpace{Test Scenes}
We ensure that the test scenes are complete, sufficiently large rooms. The following scenes are used for evaluation: 
\begin{itemize}[noitemsep,topsep=0pt,parsep=0pt,partopsep=0pt]
    \item scene0710\_00
    \item scene0758\_00
    \item scene0781\_00 
\end{itemize}

\begin{figure}[htb]
  \centering
\begin{tikzpicture}[squarednode/.style={rectangle, draw=white, fill=white, very thin, minimum size=2mm, text opacity=1,fill opacity=0,draw opacity=0}]
        \node[anchor=south west,inner sep=0] (image) at (0,0)
        {\includegraphics[width=\linewidth,trim={-0.1cm 0.3cm 0.0cm 0.2cm},clip]{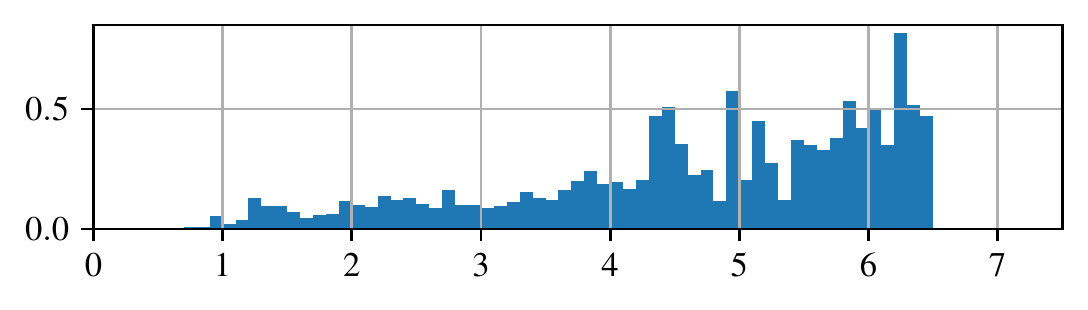}};
        \begin{scope}[x=(image.south east),y=(image.north west)]
        \footnotesize
        \node[squarednode] at (0.5, -0.125) (a) {Depth $z$ [m]};
        \node[squarednode, rotate=90] at (0.0, 0.5) (b) {MAE [m]};
        \end{scope}
\end{tikzpicture}
   \caption{SfM depth error on ScanNet.
   }
   \label{fig:sfm_error}
\end{figure}

\paragraphNoSpace{SfM Quality on Few Views}
\Cref{fig:sfm_error} shows the mean absolute error (MAE) of the SfM points against the sensor depth. It is computed on the 6291 points from the three ScanNet evaluation scenes. The maximal error is 5.85m. We do not filter the COLMAP SfM output, i.e., all points are projected to the corresponding input views and used as input to the depth completion. 

\subsection{Matterport3D \cite{Chang2017Matterport3DLF}}
\paragraphNoSpace{Train/Test Image Selection}
Similar to ScanNet, it is ensured that surfaces are observed from at least one input view. 
25\% of the train pixels are not observed by any other train view, 45\% are observed by one other, 30\% by two or more. Test views have on average 67\% overlap with their most overlapping train view. 

\paragraphNoSpace{Image Resolution}
The image resolution is 504$\times$630 after downsampling and cropping dark borders from calibration. 

\paragraphNoSpace{Test Scenes}
We avoid unbounded open space, which is challenging for NeRF approaches. The following scenes are used for evaluation: 
\begin{itemize}[noitemsep,topsep=0pt,parsep=0pt,partopsep=0pt]
    \item Region 5, house VzqfbhrpDEA
    \item Region 2, house Vvot9Ly1tCj
    \item Region 19, house Vvot9Ly1tCj
\end{itemize}

\section{Impact of Sparse Depth Density}
\label{sec:sparse_depth_density}
We investigate the impact of the sparse depth density on Matterport3D by decreasing it from 0.1\% to 0.05\% and 0.01\% (\cref{tab:results_sparse_depth_matterport}). 
\begin{table}[tb]
  \centering
\resizebox{\linewidth}{!}{
  \begin{tabular}{@{}lc@{\,\,}c@{\,\,}c@{\,\,}c@{\,\,}c@{}}
    \toprule
    & Sparse depth & & & & Depth \\
    Method & density & PSNR$\uparrow$ & SSIM$\uparrow$ & LPIPS$\downarrow$ & RMSE $\downarrow$ \\
    \midrule
    Ours w/o completion & 0.10\% & 16.90 & 0.615 & 0.521 & 0.427 \\
    Ours & 0.10\% & \textbf{18.33} & \textbf{0.673} & \textbf{0.402} & \textbf{0.114} \\
    Ours & 0.05\% & 18.10 & 0.662 & 0.414 & 0.136 \\
    Ours & 0.01\% & 17.99 & 0.662 & 0.437 & 0.140 \\
    \bottomrule
  \end{tabular}
  }
  \caption{Impact of sparse depth density on Matterport3D. Depth RMSE is in meters.}
  \label{tab:results_sparse_depth_matterport}
\end{table}
While reduced sparse depth lowers performance, it clearly shows that depth completion increases the value of very sparse depth input: With just one tenth of the sparse depth our method still performs better, than the version without completion. Despite 0.01\% being very sparse---just 32 points per image on average---we expect that using monocular depth estimation is challenging as view-consistent depth is needed. 

\section{Implementation Details}
\label{sec:implementation_details}
\subsection{Our Method}

\paragraphNoSpace{Radiance Fields}
Our model architecture is based on NeRF~\cite{Mildenhall2020NeRFRS}. The encoded position $\gamma(\mathbf{x})$ is provided as input to the first of 8 layers as well as to the fifth, by concatenating it with the activations from the fourth layer. Layers 1--8 each have 256 neurons and ReLU activations. The output of layer 8 is passed through a single layer with softplus activation to produce density $\sigma$. The output of layer 8 is also passed through a 256-channel  layer without activation, whose output is concatenated with the viewing direction $\mathbf{d}$ and the latent code $\boldsymbol{\ell}$. The concatenated vector is fed to a 128-channel layer with ReLU activation, before the final layer producing the color $\mathbf{c}$. 
The latent codes ${\boldsymbol{\ell}}$ have a size of 4 on ScanNet and 16 on Matterport3D. 
Due to the different characteristics of the depth input on the two datasets, a suitable depth loss weight $\lambda$ is determined for each approach and dataset and used across all scenes of the same dataset (\cref{tab:depth_loss_weight}). 
\begin{table}[tb]
  \centering
  \small
  \begin{tabular}{@{}lcc@{}}
    \toprule
     & ScanNet & Matterport3D \\
    \midrule
    Ours w/o Completion & 1.0 & 0.25 \\
    Ours w/o Uncertainty & 0.001 & 0.007 \\
    Ours w/o GNLL & 0.04 & 0.03 \\
    Ours w/o Latent Code & 0.003 & 0.007 \\
    Ours & 0.003 & 0.007 \\
    \bottomrule
  \end{tabular}
  \caption{Depth loss weights $\lambda$.}
  \vspace{-0.3cm}
  \label{tab:depth_loss_weight}
\end{table}

\paragraphNoSpace{Depth Completion}
The depth completion network is based on the architecture from Cheng \etal \cite{Cheng2020LearningDW}. We use a ResNet-18~\cite{He2016DeepRL} encoder and add a second upsampling branch for uncertainty estimation. It equally consists of up-projection layers with skip connections to the same downsampling layers as the depth prediction branch. To increase performance on very sparse input depth, both branches use a CSPN module, configured to 48 iterations in the depth branch and 24 iterations in the standard deviation branch. 
The depth completion network is trained at a lower resolution of 256$\times$320 on Matterport3D, and 240$\times$320 on ScanNet. 
We use the Adam optimizer~\cite{Kingma2015AdamAM} with a learning rate of 0.0001 and a batch size of 8. We train for 50 epochs on Matterport3D and 12 epochs on ScanNet. 
On Matterport3D 80 houses are used for training, 5 houses for validation, and 5 houses for testing. On ScanNet we use the provided data split. We ensure that the scenes used for NeRF are not included during training, and are instead in the test sets.

\subsection{NerfingMVS~\cite{wei2021nerfingmvs}}
\label{sec:implementation_details_nerfingmvs}
The error map calculation used by NerfingMVS was not sufficiently robust to by applied to entire rooms, so to improve this baseline's performance we adapted it as follows:

\paragraphNoSpace{Original Calculation} For each input view an error map is computed by projecting the 3D points according to the depth prior to all other views, where a depth reprojection error is computed and normalized with the projected depth. 
The mean of the 4 smallest errors are used as values in the error map.

\paragraphNoSpace{Problem on Entire Rooms} When applying the computation on entire rooms as opposed to a local region, the projected 3D points from other views frequently lie behind the camera. As a result the computed mean is often negative. 
Similarly, the computation of the near and far planes of the scenes is not suited for entire rooms, leading to a negative near plane in our case. 
Negative near plane and negative error map content lead to invalid sampling ranges, where the far bound lies in front of the near bound. to address this, we set the near and far planes ($t_n$ and $t_f$) of each scene such that all depth prior values are contained. 
In the error map calculation, we assign a maximal error $t_{f} - t_{n}$ for projected points that lie behind the camera. Afterwards, the error map values are still computed as the mean of the smallest 4 errors. 

\subsection{DS-NeRF~\cite{Deng2021DepthsupervisedNF}}
We used the same positional encoding frequencies as described for our method in the main paper for this baseline, which improved its performance.
A depth loss weight of 0.1 was suitable for the ScanNet scenes. 

\subsection{NeRF~\cite{Mildenhall2020NeRFRS}}
As in DS-NeRF, we used our own positional encoding frequencies for this baseline, which improved its performance.

\end{document}